# A Hybrid Model for Combining Neural Image Caption and k-Nearest Neighbor Approach for Image Captioning

Kartik Arora, Ajul Raj, Arun Goel, Seba Susan[0000-0002-6709-6591]*

Department of Information Technology,
Delhi Technological University,
Bawana Road, Delhi, India-110042
seba_406@yahoo.in

**Abstract.** A hybrid model is proposed that integrates two popular image captioning methods to generate a text-based summary describing the contents of the image. The two image captioning models are the Neural Image Caption (NIC) and the *k*-nearest neighbor approach. These are trained individually on the training set. We extract a set of five features, from the validation set, for evaluating the results of the two models that in turn is used to train a logistic regression classifier. The BLEU-4 scores of the two models are compared for generating the binary-value ground truth for the logistic regression classifier. For the test set, the input images are first passed separately through the two models to generate the individual captions. The five-dimensional feature set extracted from the two models is passed to the logistic regression classifier to take a decision regarding the final caption generated which is the best of two captions generated by the models. Our implementation of the *k*-nearest neighbor model achieves a BLEU-4 score of 15.95 and the NIC model achieves a BLEU-4 score of 16.01, on the benchmark Flickr8k dataset. The proposed hybrid model is able to achieve a BLEU-4 score of 18.20 proving the validity of our approach.

## 1 Introduction

Image captioning is the task of generating text that describes a given image. Describing the contents of an image in a textual way has many applications, for example, describing contents on a screen for visually impaired, real time captioning of videos, and in robotics. Image captioning is different from image classification since it involves not only identifying the objects in the image, but also summarizing the relation between the objects in the image using natural language. A lot of research work has been done in this topic in recent times [1] [2] [3] [4]. One method is to use the k-Nearest Neighbor (kNN) approach [2] to select a caption in the dataset that accurately describes the image. This involves finding a consensus caption from a set of captions that describe images that are similar to the test image. If the set of images are diverse, one would expect the selected caption to be generic (example- a dog). If the images are similar, the caption selected would be more specific (example- a black dog).

Another approach is to use the Neural Networks to generate novel captions that describe the test image. The model in [1] uses a recurrent neural network for generating the sentences and is also called Neural Image Caption (NIC). This approach uses a combination of pre-trained convolutional neural network VGG16 that processes the input image, and Long-Short Term Memory (LSTM) [5] which is well suited for processing sequential data i.e. the captions in this case. We propose to integrate the two approaches- NIC and kNN into a hybrid model that uses a trained logistic regression classifier to choose the better caption. If the test image is quite similar to the images in the training set, one would expect the captions generated by Nearest Neighbor be better than the Neural Network approach. Otherwise, the novel captions generated by NIC tend to be better. We seek to find a set of criteria to choose the model that would provide the better captions for an input image. We use Flickr8K dataset to evaluate our model. The organization of this paper is as follows: the related work is discussed in section 2, the proposed hybrid model is presented in section 3, the results are analyzed in section 4 and the conclusion is drawn in section 5.

## 2 Related Work

Image caption generation is mostly implemented either by distance-based matching or by training neural networks like LSTM. Distance or similarity based classifiers have managed to carve their own niche despite the success of neural networks for image classification [10]. This fact is reconfirmed through our own experiments which prove that for several examples, the distance-based classifier outperforms the neural network in caption generation. A hybrid model incorporating the goodness of both distance-based and neural network approaches is proposed in our work and will be described in detail in subsequent sections. We discuss works pertaining to both approaches in this section. Devlin *et al.* proposed a k-Nearest Neighbor (kNN) approach for image captioning [2]; this is the kNN model in our hybrid technique. This approach generates a caption for the test image using the captions of images in the training set that are similar to the test image. This approach finds the nearest $k$ images for the test image using the cosine similarity metric, for three different feature spaces: GIST, fc7 and fc7-fine. Each of the $k$ images have 5 captions each, so the candidate caption set C consists of $n=5*k$ captions. Then the Consensus Caption $c^*$ according to [2] is the caption with highest similarity score (BLEU-4 score [7]) with all the captions within subset M of C.

$$c^* = \underset{c \in C}{\operatorname{argmax}} \max_{M \subset C} \sum_{c' \in M} Sim(c, c') \qquad (1)$$

Vinyals *et al.* proposed a neural network model called Neural Image Caption (NIC) to generate novel image captions; this is the second model used in our hybrid technique. It consists of an encoder CNN connected to an LSTM network. The CNN is pre-trained on image classification task and the last hidden layer of CNN is used as input to the LSTM network. The model maximized the probability $p(S|I)$ where $I$ is the image and $S$ is a sequence of words $\{S_1, S_2, ...\}$ that describes the image. The model used the CNN to extract a feature vector from the input image which is then used as an input to the

LSTM circuit. Then, using the encoded image and the partial caption (which at the beginning would be null or a special start token), the output of the LSTM would be the probability of each word in the dictionary to be the next word in the sequence, out of which we either take the one with the maximum probability (greedy search), or choose the top *i* words (beam search). The word would be added to the previous partial caption to generate the new partial caption. The caption would end when a special end token was selected or a specified length was reached. The model was trained using stochastic gradient descent [8] minimizing the loss function:

$$L(I, S) = -\sum_t \log p_t(S_t) \qquad (2)$$

The loss function is the summation of negative log probabilities of correct word $S_t$ at each step *t*. Before training, basic pre-processing is done on captions in the dataset. All words with occurrences greater than 5 are kept in the dictionary. Unlike the k-Nearest Neighbor (kNN) model which chooses a caption from the training set that best represents the test image, NIC Generator can construct novel captions which are not present in the training set. Recent literature focusses on modifying LSTM-based network architectures for improvising the natural language in image captions. A hierarchical LSTM in a recent work [9] comprises of a phrase decoder and a sentence decoder for generating image description. Xu *et al.* [4] developed an attention-based model which was able to focus on relevant parts of the image to generate better captions. Ding *et al.* [3] proposed that instead of assigning equal weights to all words, one could assign different weights to words according to their importance in the sentence. For example, for the images of a bird, the word 'bird' would have a larger weight. This also prevents mis-recognition since the main subjects in the image are identified correctly.

## 3 Proposed Hybrid Model

In this section, we propose a hybrid model that combines two state-of-the-art models: Neural Image Caption (NIC) [1] and *k*-Nearest Neighbor approach [2] to generate captions for an input image. Both models were described in detail in section 2. In our NIC implementation, the image is first fed to a pre-trained convolutional neural network, Inception-V3 [6], that produces a rich representation of the input image by encoding it into a fixed-length vector of size 2048. This vector is the output of the last hidden layer of the Inception-V3 model and it is given as input to a LSTM which is a recurrent neural network.

Our hybrid technique incorporates a meta-classifier (logistic regression) that will choose the better model for a given input image and use the caption generated by this model. The general idea for a hybrid model is shown in Fig. 1. We propose a generic algorithm which requires classifying an image into either category A or category B, where category A is the category of images that are better modeled by NIC and category B is the category of images that are better modeled by k-Nearest Neighbor model. We use logistic regression for this classification and discuss some possible set of features which can help us to produce a robust classifier.

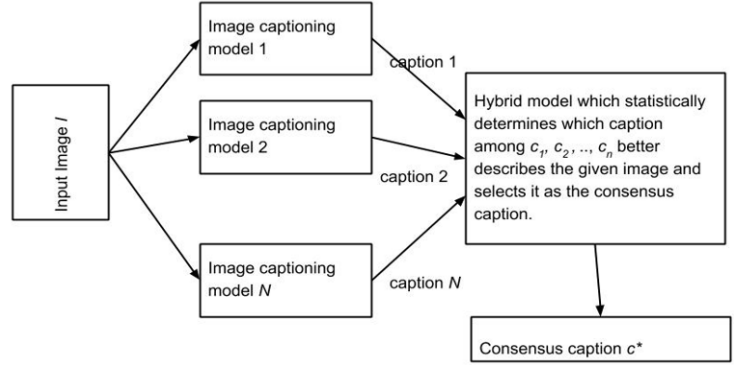

**Fig. 1.** Generic layout of a hybrid model which combines *N* image captioning models

### 3.1 Methodology

Let $M_1$ and $M_2$ be the NIC model [1] and the *k*-nearest neighbor model [2], respectively that are trained individually on the training set of images and their ground-truths (captions). We propose a hybrid model *M* which selects a consensus caption *c\** for a given input image *I* using the following steps.
1. Generate caption $c_1$ using $M_1$ for given input image *I*.
2. Generate caption $c_2$ using $M_2$ for given input image *I*.
3. A set of five features (section 3.2) are extracted from the two models using the validation set and fed as input to the logistic regression classifier.
4. The BLEU-4 scores of the two models are compared for generating the binary-value ground truth for the logistic regression classifier (0 if BLEU($M_1$)≥ BLEU($M_2$), else 1 if BLEU($M_1$)< BLEU($M_2$)).
5. For the test set, the input images are first passed separately through the two models to generate the individual captions. The five-dimensional feature set extracted from the two models is passed to the logistic regression classifier to take a decision regarding the final caption generated which is the best of two captions generated by the models.
6. If the logistic regression classifier predicts that $M_1$ produces a better caption i.e. predicted value *y*=0 for *I*, then *c\** = $c_1$, else if *y*=1, then *c\** = $c_2$.

The block diagram for the process is shown in Fig. 2.

### 3.2 Feature extraction and normalization

We propose a set of five features extracted from the classification results of models $M_1$ and $M_2$ that is used for training the meta-classifier in our hybrid model. The qualitative definitions of the new features are enlisted as follows.
   a. The confidence score that the NIC model has for the caption it generated for the given image.

b. The confidence score that the $k$-nearest neighbor model has for the caption it generated for the given image.
c. A measure of similarity between the images in the training data to the input image in consideration.
d. The length of the captions generated by both the models.

The above conditions led us to formulate the five features quantitatively as follows.
1. Length-normalized log probability $p^*$ of $c_1$ (from $M_1$) which is a measure of the confidence $M_1$ has on $c_1$.
2. The average (BLEU-4) similarity score of $c^*$ (from $M_2$), from (1).
3. Cosine similarity $S_c$ between input image $I$ and the image $Y$, where $Y$ belongs to $K$ the set of the $k$ nearest images to $I$, summed over all $Y$. Averaging the similarity scores across multiple samples of a class improves accuracy as observed in [12]. The features are derived from the fc7 layer of VGG16 pretrained network used in model $M_2$ [2].

$$S_c = \sum_{Y \in K} \cos\_sim(I, Y) \qquad (3)$$

4. The two features: length of caption $c_1$ ($=l_1$) and length of caption $c_2$ ($=l_2$).

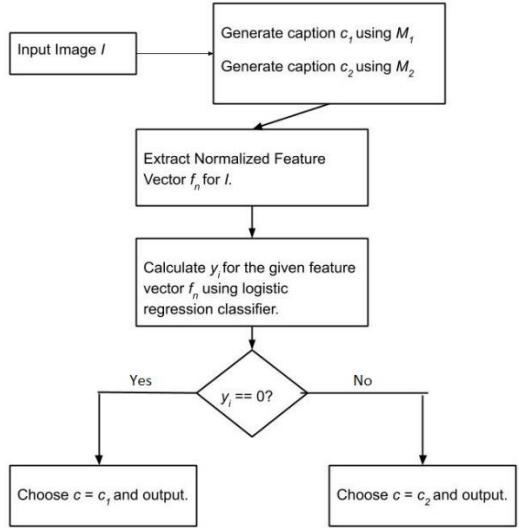

Fig. 2. Process flow of generating consensus caption $c^*$ for input image using proposed model $M$

These features are now normalized so that their absolute values lie in the range [0, 1]. We divide both $l_1$ and $l_2$ by the length of the longest caption in the Flickr8k dataset (=35 words). We divide $S_c$ by five times the number of summands in its summation (i.e. $5k$) in (3) and we divide $\sum_{c' \in M} Sim(c^*, c')$ by the number of summands in its summation (i.e. $|M|$) in (1). Finally, we have the normalized feature vector $f_n$ given by (4) that is the input to the logistic regression classifier.

$$f_n = \left\{ p^*, \frac{\sum_{c' \in M} Sim(c^*, c')}{|M|}, \frac{S_c}{5k}, \frac{l_1}{35}, \frac{l_2}{35} \right\} \quad (4)$$

## 4 Results

We use Flickr8K dataset [11] to evaluate our hybrid model. It contains 8092 images with 5 captions each, out of which 6000 are used for training, 1000 for testing and the rest for development. We first compare BLEU-1 and BLEU-4 scores for various LSTM beam sizes in Table 1. In Table 2, we present results for: 1) Neural network based NIC model [1], with beam size $i = 3$ which has the highest BLEU-1 score in Table 1, 2) $k$-nearest neighbor model [2] with $k = 30$ and $|M| = 50$, 3) Proposed hybrid model which integrates the above two models using the logistic regression classifier. All the scores reported have been evaluated on the Flickr8k dataset on a system with Intel® Core™ i5-8300H Processor, with 8 GB RAM and GTX 1050 graphics running on Windows 10 Pro 64 bit. The code was compiled on Python 3.6.9 using TensorFlow 2.1.0. The proposed hybrid model was able to achieve higher BLEU-1 and BLUE-4 scores on the Flickr8k test data than the individual models as observed from Table 2.

**Table 1**. BLEU-1 and BLEU-4 scores for NIC for different beam sizes.

| Beam Size ($i$) | BLEU-1 | BLEU-4 |
|---|---|---|
| 1 | 57.59 | 14.44 |
| 3 | 58.13 | 16.01 |
| 5 | 58.09 | 16.29 |
| 7 | 57.89 | 16.03 |

**Table 2**. BLEU-1 and BLEU-4 scores for kNN, NIC and Hybrid Model.

| Model | BLEU-1 | BLEU-4 |
|---|---|---|
| kNN with $k = 30$, $M = 50$ | 56.02 | 15.95 |
| NIC with beam size 3 | 58.12 | 16.01 |
| Hybrid model | 59.67 | 18.20 |

Table 3 shows some examples of captions generated using our hybrid scheme. One of the captions (either NIC or kNN) shown in each of the five cases is incorrect. As observed, in two cases out of five, the kNN model outperforms the neural network approach (NIC). Our hybrid model chooses the best caption that describes the scene adequately in all five cases. The code of our hybrid model is made available online at https://github.com/rizal-rovins/hybrid-image-captioning-model

**Table 3.** Images and their captions generated by the hybrid model.

| | | | |
|---|---|---|---|
| 1 | 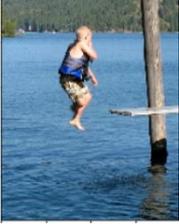 | NIC Caption | Hybrid model classifies image to category A (NIC). *Final caption:* A boy is jumping off a dock into a lake. |
| | | A boy is jumping off a dock into a lake. | |
| | | kNN Caption | |
| | | A woman in a bikini jumping off a dock into a lake. | |
| 2 | 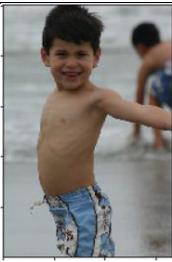 | NIC Caption | Hybrid model classifies image to category B (kNN). *Final caption:* A boy jumping in the air on the beach. |
| | | A little girl in pink bathing suit is jumping into the water. | |
| | | kNN Caption | |
| | | A boy jumping in the air on the beach. | |
| 3 | 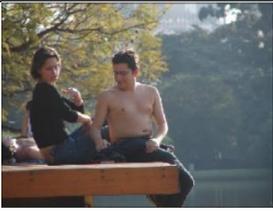 | NIC Caption | Hybrid model classifies image to category A (NIC). *Final caption:* A man and a woman are sitting on a park bench. |
| | | A man and a woman are sitting on a park bench. | |
| | | kNN Caption | |
| | | A girl doing a handstand on a trampoline. | |
| 4 | 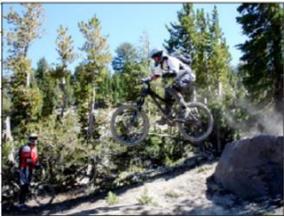 | NIC Caption | Hybrid model classifies image to category A (NIC). *Final caption:* A mountain biker rides through the woods. |
| | | A mountain biker rides through the woods. | |
| | | kNN Caption | |
| | | A man riding a bike down a hill. | |
| 5 | 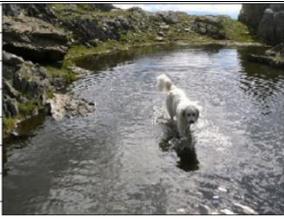 | NIC Caption | Hybrid model classifies image to category B (kNN). *Final caption:* A dog running through the water. |
| | | A white dog fetches a stick in his mouth. | |
| | | kNN Caption | |
| | | A dog running through the water. | |

## 5   Conclusion

We have presented a hybrid model that combines two existing image captioning models- NIC and k-Nearest Neighbor (kNN) trained separately on images from the training set. We extract a novel set of five features from the validation set for evaluating the captions generated by the two models, that is used to train a logistic regression classifier. The BLEU-4 scores of the two models are compared for generating the {0, 1} ground truth values for the logistic regression classifier. Our hybrid model chooses the best caption that describes the scene adequately for a given test image. The proposed method was able to achieve higher BLEU-1 and BLUE-4 scores on the benchmark Flickr8k dataset. The technique can be further extended to combine more than two image captioning models and advanced forms of LSTM incorporating attentional mechanism could be used in place of NIC in our model.